\documentclass[letterpaper]{article}
\usepackage{aaai25}
\usepackage{times}
\usepackage{helvet}
\usepackage{courier}
\usepackage[hyphens]{url}
\usepackage{graphicx}
\usepackage{natbib}
\usepackage{caption}
\usepackage{amsmath,amssymb}
\usepackage{algorithm}
\usepackage{algorithmic}
\usepackage{xcolor}
\copyrightyear{2026}
\nocopyright  

\newif\ifshowchanges
\showchangesfalse
\ifshowchanges
  \newcommand{\added}[1]{{\color{blue}#1}}
  \newenvironment{addedtext}{\color{blue}}{}
  \newcommand{\todo}[1]{{\color{red}[TODO: #1]}}
\else
  \newcommand{\added}[1]{#1}
  \newenvironment{addedtext}{}{}
  \newcommand{\todo}[1]{}
\fi

\title{After 15 Years of a Sleeping Beauty: Revitalizing and Extending the GPU Planner}
\author{
  Andrew Dougherty\textsuperscript{\rm 1}\qquad Stefan Edelkamp\textsuperscript{\rm 2}
}
\affiliations{
  \textsuperscript{\rm 1}Independent Researcher, Michigan, USA\\
  andrew.dougherty.research@gmail.com\\
  \textsuperscript{\rm 2}Department of Theoretical Computer Science and Mathematical Logic,\\
  Faculty of Mathematics and Physics, Charles University, Prague\\
  edelkamp@ktiml.mff.cuni.cz
}

\begin{document}
\maketitle

\begin{abstract}
CuDPlan was an early GPU planner for cost-optimal classical planning.
It combined a compact postfix representation of grounded actions with batched
state expansion on the graphics card and hash-based duplicate detection on the
CPU.
After a long period in which the code no longer compiled on current hardware,
we revived it and extended it with a new family of finite-domain potential
heuristics.  The extension is deliberately non-invasive: the original atomic
sidecar format and its default behaviour are preserved, while a versioned
sidecar carries conjunctions of state-bit literals and two integer weights per
feature.  Offline, a mixed-integer linear program selects unary and pairwise
features and their weights; online, the GPU evaluates the resulting potential
in the existing A$^\ast$ loop.
\added{On the 1{,}827 tasks of a recent GPU-planning benchmark suite, the
revived planner with unary potentials solves 668 tasks in five minutes, more
than the published results of a GPU-accelerated $h^2$ planner in thirty
(obtained on different hardware); a portfolio that runs
it for five minutes and then hands over to Fast Downward with LM-cut solves
1{,}003 tasks within thirty minutes, against 968 for LM-cut alone.}
\end{abstract}

\section{Introduction}
\added{GPU planning has a long but uneven history.}  Around 2010,
CuDMoC/CuDPlan (henceforth CuDPlan) explored how much of an optimal state-space search could be
moved to a graphics card: states were expanded in batches on the GPU, while
duplicate detection used a lock-free CPU hash table, and the transition relation
was represented by compact postfix programs
\citep{edelkamp2010perfect,sulewski2011exploiting}.  The
approach was attractive because state-space search is regular, branch-poor, and
dominated by memory traffic---properties that fit SIMD hardware well
\citep{edelkamp2012heuristic}.

The code, however, aged.  It relied on CUDA idioms and constant-memory tables
that current compilers reject, and on host-side writes to device symbols that
were never well-defined.  Simply compiling it again was therefore not enough.
We rebuilt the relevant pieces for a modern toolchain, fixed state-bit updates
that corrupted bits when an effect re-asserted an already true fact, and moved
large tables out of the 64 KB constant-memory limit.  The revived planner runs
on most current GPUs including Metal and NVIDIA architectures, and retains the
original cost-optimal A$^\ast$ search.  The hardware case has only
strengthened: in modern workloads and synthetic benchmarks, the RTX 5090
calculates roughly 90 to 100 times faster than the original GTX 480 on
paper---a
9{,}000\%+ performance leap (Table~\ref{tab:gpus})---and 50{,}000-core GPUs
have already been announced \citep{nvidia_rubin_hpc_2026}.

\begin{table}[t]
\centering
\footnotesize
\begin{tabular}{lrr}
\hline
& \textbf{GTX 480 (2010)} & \textbf{RTX 5090 (2025)} \\
\hline
Architecture & Fermi (GF100) & Blackwell (GB202) \\
Process node & 40 nm & 5 nm (TSMC 4N) \\
Transistors & 3.1 billion & 92.2 billion \\
CUDA cores & 480 & 21{,}760 \\
Video memory & 1.5 GB GDDR5 & 32 GB GDDR7 \\
Memory bandwidth & 177.4 GB/s & 1{,}792 GB/s \\
Power (TDP) & 250 W & 575 W \\
\hline
\end{tabular}
\caption{The original target hardware and a current card.}
\label{tab:gpus}
\end{table}

The scientific question is whether the revived engine can support a stronger
heuristic without sacrificing its simple search architecture.  We answer with
\emph{finite-domain potential heuristics} \citep{pommerening2014lp,
pommerening2017higher}.  A potential assigns weights to local features of a
state and sums them; consistency with action costs guarantees admissibility.
The original planner already had an atomic potential: one bit, one true weight,
one false weight.  We add conjunctions of literals, in particular pairs of
variable-value assignments.  The extension is off by default: version-1
sidecars keep their old semantics, and the new version-2 sidecar is activated
only when its header is present.

Our contributions are:
(1) a modern, working port of CuDPlan with correctness fixes;
(2) a backward-compatible GPU sidecar for conjunctive finite-domain
potentials and an offline MILP generator;
(3) a controlled evaluation of unary versus pairwise features on
\added{fifteen} IPC domains, including the large barman instance;
(4) evidence that pairwise features can be highly effective, but are not
monotone: they lose on blocks, freecell, and scanalyzer even when their initial
state value is no smaller\added{;
(5) an independent reproduction of these results on a second, current GPU;
(6) a full-suite coverage comparison with a recent GPU-accelerated $h^2$
planner \citep{fritzsche2026gpu}, including runs with a larger closed list;
and (7) a simple portfolio with Fast Downward's LM-cut that solves more tasks
than either component}.

\section{Related Work}
Potential heuristics were introduced as a compact alternative to delete-relaxation
and pattern-database heuristics \added{\citep{pommerening2015nonnegative}}.  The LP-based construction of
\citet{pommerening2014lp} and the higher-dimensional generalization of
\citet{pommerening2017higher} are the closest relatives of our feature family;
\citet{seipp2015new} study alternative objectives for the same LP.  Our work
differs in implementation: the potentials are compiled into a versioned
sidecar and evaluated inside a GPU A$^\ast$ engine, with no change to the
search code.

The GPU planning line goes back to perfect hashing and batched expansion
\citep{edelkamp2010perfect,sulewski2011exploiting}.  The revived code keeps
their state-space organization but modernizes memory placement and fixes
state-update correctness.  The default closed list is the lock-free CPU hash
table; hashing on the GPU remains an experimental compile-time option.
\added{\citet{fritzsche2026gpu} take the opposite division of labour, computing
$h^2$ on the GPU inside a CPU search; batching an $h^2$-style heuristic inside
CuDPlan's GPU loop, which already evaluates potentials in batches, would combine
the two.}  On the satisficing side, \citet{kuroiwa2018batch} batch Monte-Carlo
random walks on the GPU and reach competitive coverage; like our A$^\ast$ they
evaluate many states per kernel launch, but target the satisficing rather than
the cost-optimal setting.  The explicit-state model-checking line shares the
same batched architecture: GPUexplore \citep{wijs2014gpuexplore,wijs2016gpuexplore2}
performs BFS-based reachability on the GPU, extended to partial-order reduction
\citep{neele2016partial}, linear-time property checking \citep{wijs2016bfs}, and
probabilistic state-space construction \citep{heemstra2025gpuexploresc};
\citet{wijs2016manycore} survey the many-core design space.  Fast Downward's
translator supplies the SAS$^+$ representation \citep{helmert2006fast}; our
binary encoding follows standard SAS$^+$ practice.  Pattern databases
\citep{haslum2007domain} and landmark heuristics remain orthogonal; the sidecar
interface could carry them, but the present work is restricted to potentials.

\section{The Revived Planner}
\subsection{Grounded model and search}
The planner reads a grounded PDDL model in the syntax accepted by
\emph{gdl-parser}.  Actions are compiled into a postfix representation of
preconditions and effects, and A$^\ast$ \citep{hart1968astar} expands states in GPU batches.  The
closed list is a lock-free hash table maintained by eight CPU threads, derived
from the shared hash table of LTSmin \citep{laarman2010boosting}; the open
list stores states with their $g$-values and heuristic estimates.  The GPU
produces successors and heuristic values, and the CPU removes duplicates before
the next batch.  With a zero heuristic the engine is a
uniform-cost search and therefore returns a cost-optimal plan.  With a
consistent potential it remains optimal.

The 2026 port keeps this design but replaces the fragile parts.  Postfix
tables now live in ordinary global memory, allocated at their exact size and
cached through L1/L2, instead of a 25 MB constant array.  Device pointers are
uploaded with \emph{cudaMemcpyToSymbol} rather than written from host code.
Porting the code with modern tooling also surfaced two latent defects that
can affect optimality in rare cases: an additive bit update that could
carry into a neighbouring bit when a true fact was re-asserted, and an
unsynchronised cost update between duplicate-detection threads.  Both are
fixed; ThreadSanitizer and systematic plan validation, which were not
standard practice in 2010, made them visible.
These changes are independent of the heuristic extension.
\added{Because the original PDDL grounder was missing from the
recovered sources, the port includes a new grounder, plan output, and a test
suite that checks every plan with the independent validator VAL
\citep{howey2004val}.}

\subsection{Two front ends and a shared numeric core}
The planning model is not lowered to DVE.  Instead, \emph{dve-parser} and
\emph{gdl-parser} are two front ends to the same model object and postfix
interpreter.  The DVE front end supports byte and integer state variables and
arrays; the postfix evaluator has distinct opcodes for scalar byte and integer
variables and for byte and integer arrays.  The current PDDL front end is
propositional: the grounder emits one Boolean atom per fact, or a binary
SAS$^+$ bit, plus an integer cost field, and the version-2 features in this
paper range over those bits.  Consequently the implementation is not yet a
numeric planner, but its state layout and kernels already accommodate numerical
state.  Extending the grounder and the potential LP to numeric variables is a
concrete path beyond the Boolean fragment, not a claim that numeric PDDL is
supported today.

\subsection{Portable kernels}
The five GPU kernels---hashing, guard checks, effect application, and metric
and heuristic evaluation---now live in a single Slang source, compiled by
\emph{slangc} to CUDA/PTX for NVIDIA and to Metal MSL for Apple GPUs; only
the host glue is per-platform (the CUDA driver API on one side, a raw Metal
host on the other).  On Metal the state buffers sit in unified memory, so no
explicit device copies remain in the hot path.  Every kernel's output was
byte-compared against the CUDA reference, and the Slang path solves the
barman case with the same cost and state count as the CUDA path (the
version-2 sidecar currently runs on the CUDA path only, as noted below).
The planner builds unmodified on Linux with the NVIDIA toolchain, so it is
directly deployable to GPU cloud providers.

\section{Finite-Domain Potential Heuristics}
\subsection{Potentials and consistency}
Let $s$ be a state and let $\Phi=\{\varphi_1,\ldots,\varphi_m\}$ be local
features, each a conjunction of signed state-bit literals.  A feature has two
integer weights, $w_i^{+}$ and $w_i^{-}$, and contributes $w_i^{+}$ if
$\varphi_i$ holds in $s$ and $w_i^{-}$ otherwise.  The potential is
\begin{equation}
h(s)=\sum_{i=1}^{m}\left[w_i^{+}\,\mathbf{1}(\varphi_i\subseteq s)+
w_i^{-}\,\bigl(1-\mathbf{1}(\varphi_i\subseteq s)\bigr)\right].
\label{eq:potential}
\end{equation}
It is consistent if for every applicable action $a$ with cost $c(a)$ and every
state $s$ satisfying its preconditions,
\begin{equation}
h(\mathrm{apply}(s,a))-h(s)\;\le\;c(a).
\label{eq:consistency}
\end{equation}
Consistency implies admissibility, so A$^\ast$ with $h$ is optimal.
The evaluation is a natural GPU workload: it is memory-bound and
broadcast-shaped.  Each thread streams its own state but reads the identical
small weight tables, which stay resident in the read-only caches; the
per-state compute is a check-and-add, with no inter-thread communication,
no atomics, and uniform control flow, since every thread evaluates every
feature.  The hardware has grown into this pattern since the planner was
written: the Kepler generation added the dedicated read-only data cache
with the \emph{\_\_ldg} intrinsic for plain global loads, and the modern
architectures unify the L1 with the texture path and carry multi-megabyte
L2 caches, so the precomputed potential tables are served as broadcasts at
the near-full bandwidth---the precomputed heuristic is exactly the workload
these developments favor.

For a unary feature $x=v$ this is the usual atomic potential.  For a pairwise
feature $(x=v,y=w)$ it captures positive or negative interaction that a sum of
unary terms cannot express.  The price is a larger feature set and a harder
offline optimization problem.

\subsection{Offline MILP}
We generate candidate features from the SAS$^+$ task: all unary assignments,
all pairs of variables that occur together in an operator precondition or
effect, and pairs formed by a goal variable and each other variable.  For each
feature we create two weight variables and linearize the change in
\eqref{eq:potential} with non-negative auxiliary variables.  The resulting
program maximizes $h(s_0)$ subject to \eqref{eq:consistency} for every
operator, with $-1000\le w_i^{\pm}\le1000$ and integer weights.  Auxiliary
variables are continuous; integer weights are sufficient for an integer-valued
sidecar and avoid numerical drift in the GPU kernel.  We solve the MILP with
HiGHS through SciPy.  The largest case reported here, barman, has 3{,}609
candidate features, 128{,}420 auxiliary variables, and 548{,}687 nonzeros; it
solves in 11.1 s.  Feature generation and MILP solving are offline and do not
enter the planner's runtime.

\subsection{Sidecars}
A sidecar is an auxiliary file that rides alongside the planner's model: it
carries the potential features and their weights, so the heuristic arrives as
data and the search code never changes.  Version 1 of the sidecar stores, for
each feature, one state bit and two weights.  Version 2 generalizes this to a variable-length conjunction.  The
file begins with \emph{CUDMOC\_POTENTIAL 2}, the feature count, and one line
per feature:
\begin{center}
\emph{arity} \ $\pm$\emph{bit}$_1$ $\cdots$ $\pm$\emph{bit}$_k$
\ \emph{wt} \ \emph{wf}.
\end{center}
A leading plus requires the bit to be true; a minus requires it to be false.
The two trailing integers are the weights when the conjunction is true and
false, respectively.  Zero-weight features are omitted.  The atomic format is
still version 1, with its original one-bit interpretation.

The loader keeps both representations in the model object.  In version 2 it
builds a compact offset array, a literal array, and a sign array, and uploads
them once to the device.  The kernel evaluates every feature for the state and
sums the selected weight.  Feature tables are read-only and benefit from the
GPU's read-only cache; the per-state work is a short loop over the literals of
each feature.  The existing expansion, hashing, and duplicate-detection code is
unchanged, so the heuristic can be enabled with the single environment variable
\emph{CUDMOC\_POTENTIAL}.  Setting it to \emph{0} disables the heuristic
and reproduces the blind baseline.  Version-1 tests and the atomic path remain
available, and the default when no sidecar is present is still no heuristic.

\subsubsection{Kernel evaluation and memory}
Algorithm~\ref{alg:eval} gives the per-state evaluation performed by the GPU
kernel.  The outer loop walks the sidecar features; the inner loop checks the
signed literals.  A state bit is addressed by byte and bit offset from the map
produced by the encoder, so the same kernel serves one-hot and binary encodings.
The two weights are stored next to the feature, and the branch on
\emph{ok} selects one of them.  The feature table is read-only, and the
offset array is small enough to stay in cache even for the largest sidecar in
this study.  The kernel computes one heuristic value per state; the existing
batched expansion and hash-table code is unchanged.

\begin{algorithm}[t]
\caption{GPU evaluation of a version-2 potential.}
\label{alg:eval}
\begin{algorithmic}[1]
\small
\STATE $h \leftarrow 0$
\FOR{$i=0$ \TO $m-1$}
  \STATE $ok \leftarrow \mbox{true}$
  \FOR{$j=\mathit{start}[i]$ \TO $\mathit{start}[i+1]-1$}
    \STATE $b \leftarrow \mathit{bits}[j]$
    \STATE $v \leftarrow (s[b\gg3] \gg (7-(b\,\&\,7))) \,\&\, 1$
    \STATE $ok \leftarrow ok \land (v \oplus \neg\mathit{positive}[j])$
  \ENDFOR
  \STATE $h \leftarrow h + (ok \,?\, \mathit{wt}[i] : \mathit{wf}[i])$
\ENDFOR
\STATE \RETURN $h$
\end{algorithmic}
\end{algorithm}

The representation is compact.  Across the pairwise runs, sidecars contain
40--2{,}136 nonzero features, 147--15{,}072 literals, and 2--193 KB of data.
Barman, the largest search but not the largest sidecar, uses 466 features,
1{,}021 literals, and 18 KB.  The 193 KB elevators sidecar is still small
compared with the search tables, which are allocated for tens of millions of
states.  The offline MILP, not the sidecar, is the expensive part: the
freecell solve takes 61.5 s, while the search itself generates only 393 states.

\subsubsection{Compatibility and fallback}
The loader accepts version 1 and version 2.  Version 1 retains the original
one-bit layout and semantics, so existing atomic sidecars and regression tests
are unaffected.  Version 2 is selected only by the header, and the old
\emph{CUDMOC\_POTENTIAL} environment variable remains the single switch.
If the file is absent or set to \emph{0}, the planner runs with $h=0$ and
returns the same cost as the original planner.  A malformed sidecar is rejected
at load time rather than silently producing an inadmissible heuristic.
Version 2 is currently implemented in the CUDA and emulation paths; the Slang
path still accepts version 1 only.  Porting it is mostly plumbing: the Slang
version-1 kernel already mirrors the CUDA control flow, and version 2 adds
three read-only feature tables, two scalar bounds, a signed-literal test, and
the host-side sidecar parser.  This is a deliberate compatibility boundary, not
a change in the default configuration.

\section{Experiments}
Experiments ran on three machines: an NVIDIA GeForce RTX 3090 (24 GB, sm\_86)
with CUDA 11.5, hosted on an AMD Ryzen 9 5900X, where the planner was built by
\emph{make gdl-parser -j4 ARCHS=86}; an NVIDIA GeForce RTX 5090 (32 GB,
power limit 400 W) hosted by an AMD Ryzen 9 9950X3D with 64 GB of RAM,
Debian 13, and Fast Downward \citep{helmert2006fast} at commit 487e33c; and
an Apple M4 Pro with 24 GB of unified memory and a 16-core GPU running
macOS 15.  The
Table~\ref{tab:results} experiments ran on the 3090; the further experiments
ran on the 5090, except the transfer results, which ran on the M4 Pro.

\subsection{Setup}
We used the IPC benchmarks bundled with
the planner and the corresponding PDDL instances from the downward-benchmarks
collection.  Each PDDL instance was translated to SAS$^+$ with Fast Downward's
translator and encoded in binary; the same encoding and the same grounded GDL
file were used for blind, unary, and pairwise runs.  Costs are action costs from
the SAS$^+$ task.  We report unique states generated by A$^\ast$, the initial
heuristic value $h(s_0)$, and the offline MILP time.  Every plan was checked
with the independent validator VAL \citep{howey2004val}, and every cost was
compared across all optimal planners.  \added{Source code, generators, sidecars, and logs will be
made available in a public repository upon publication.}

A unary sidecar uses all variable-value features of the SAS$^+$ task.  A
pairwise sidecar adds the causal and goal pairs described above.  This is the
natural 1D-versus-2D comparison for finite-domain potentials; it avoids
conflating the result with the binary-versus-one-hot encoding difference that
affects atomic bit potentials.

\subsection{Initial Results}
\begin{figure}[!htbp]
\centering
\includegraphics[width=\columnwidth]{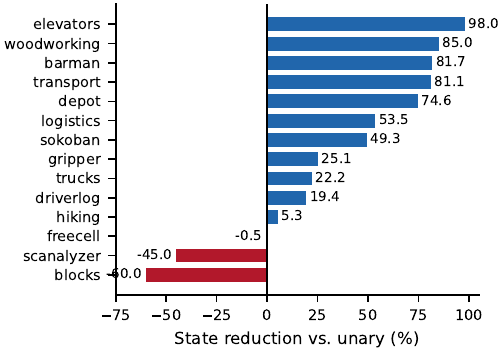}
\caption{State reduction of pairwise over unary features.  Positive values are
improvements; blocks, freecell, and scanalyzer are worse with the larger
feature set.  Tidybot is omitted because its pairwise MILP did not complete.}
\label{fig:savings}
\end{figure}

Table~\ref{tab:results} and Figure~\ref{fig:savings} summarize the results.
The barman instance is the clear success: pairwise features reduce generated
states from 5{,}645{,}887 to 1{,}035{,}201, or 81.7\% relative to unary and
83.2\% relative to blind search.  Search time falls from 8.4 s to about 3.3 s
(median of three runs; both include about 0.3 s of GPU initialization).  The initial value
rises from $h(s_0)=36$ to $h(s_0)=73$; the blind optimum is 90.

On the smaller IPC instances, pairwise features are not uniformly better.
They win on ten of the thirteen tasks for which the MILP completed, often
substantially: elevators (418{,}200 to 8{,}560 states, 98.0\%), transport
(127 to 24, 81.1\%), woodworking (4{,}481 to 671, 85.0\%), and logistics
(3{,}661 to 1{,}701, 53.5\%).  They lose on blocks (30 to 48), freecell
(391 to 393), and scanalyzer (8{,}663 to 12{,}559).  The losses are modest in
absolute terms for blocks and freecell, but large for scanalyzer: there the
initial value is unchanged at 18, while the extra features make the heuristic
less informative along the search frontier.
\added{The blocks result depends on the MILP solution: the solver can return
different weightings with the same $h(s_0)$, and in our reproduction
(below) the pairwise sidecar generated 30 states, the same as unary.}

The aggregate picture depends on the mix.  Over the thirteen non-barman
instances with pairwise results, blind search generates \added{505{,}388}
states, unary \added{439{,}582}, and pairwise 26{,}729: 93.9\% fewer than unary
and 94.7\% fewer than blind.  Including barman, the totals are
\added{6{,}663{,}718, 6{,}085{,}469}, and 1{,}061{,}930
(\added{82.5\% and 84.1\%}).  The aggregate is dominated by elevators and
barman, so the per-instance pattern is more informative than the totals.
Tidybot's pairwise MILP has 5.58 million auxiliary variables and did not finish
within 600 s; its unary run reduces 115 states to 61.

\begin{table*}[!htbp]
\centering
\footnotesize
\setlength{\tabcolsep}{3.0pt}
\begin{tabular}{lrrrrrrrrrr}
\hline
Domain & $|V|$ & $|O|$ & Blind & Unary & Pairwise & $\Delta_U$ & $\Delta_B$ & $h_U$ & $h_P$ & $t_P$ \\
\hline
barman       & 62 & 358 & 6{,}158{,}330 & 5{,}645{,}887 & 1{,}035{,}201 & 81.7 & 83.2 & 36 & 73 & 11.1 \\
blocks       & 9  & 32  & 125     & 30     & 48      & $-60.0$ & 61.6 & 6  & 6  & 0.06 \\
depot        & 14 & 72  & 487     & 413    & 105     & 74.6 & 78.4 & 4  & 10 & 0.25 \\
driverlog    & 8  & 88  & 302     & 165    & 133     & 19.4 & 56.0 & 3  & 5  & 0.07 \\
elevators    & 9  & 362 & 418{,}200 & 418{,}200 & 8{,}560 & 98.0 & 98.0 & 0 & 50 & 2.79 \\
freecell     & 22 & 504 & 4{,}287 & 391    & 393     & $-0.5$ & 90.8 & 8  & 8  & 61.5 \\
gripper      & 7  & 34  & 255     & 255    & 191     & 25.1 & 25.1 & 8  & 10 & 0.16 \\
hiking       & 7  & 110 & 625     & 589    & 558     & 5.3  & 10.7 & 2  & 5  & 1.16 \\
logistics    & 7  & 54  & 13{,}746 & 3{,}661 & 1{,}701 & 53.5 & 87.6 & 16 & 20 & 0.04 \\
scanalyzer   & 12 & 540 & 46{,}080 & 8{,}663 & 12{,}559 & $-45.0$ & 72.7 & 18 & 18 & 19.8 \\
sokoban      & 28 & 114 & 1{,}766 & 893    & 453     & 49.3 & \added{74.3} & 4  & 11 & 1.14 \\
tidybot      & 137 & 4{,}591 & 115 & 61 & --- & --- & --- & 4 & --- & $>$600 \\
transport    & 6  & 104 & 127     & 127    & 24      & 81.1 & 81.1 & 4  & 54 & 0.06 \\
trucks       & 10 & 261 & 5{,}245 & 1{,}714 & \added{1{,}333}  & 22.2 & 74.6 & 9  & 11 & 2.24 \\
woodworking  & 22 & 192 & 14{,}143 & 4{,}481 & 671    & 85.0 & 95.3 & 130 & 170 & 0.92 \\
\hline
Total (13)   &    &     & \added{505{,}388} & \added{439{,}582} & 26{,}729 & 93.9 & 94.7 & & & \\
\hline
\end{tabular}
\caption{Binary SAS$^+$ encoding.  ``Blind'' is A$^\ast$ with $h=0$;
``Unary'' and ``Pairwise'' are finite-domain potentials; the three columns
report generated states.  $\Delta_U$ and
$\Delta_B$ are percentage reductions in generated states relative to unary and
blind search.  $h_U$ and $h_P$ are $h(s_0)$; $t_P$ is the offline MILP time in
seconds.  The total excludes barman and tidybot (13 instances).  Barman costs 90 in all three configurations; all
pairwise plans were validated.}
\label{tab:results}
\end{table*}

Adding features enlarges the feasible region of the MILP but does not imply
pointwise dominance.  The objective only maximizes $h(s_0)$; a solution with a
larger initial value can still be less accurate on the states that A$^\ast$
actually visits.  The three losses illustrate the effect: blocks and freecell
already have a strong unary potential and the pairwise optimum chooses a
different, less useful distribution; scanalyzer keeps the same initial value
but loses guidance elsewhere.  This is not a bug in the sidecar or in
consistency: all weights are integer, all runs preserve the blind cost, and the
validator accepts every plan.  It is a reminder that ``higher-dimensional'' is
a feature-space statement, not a guarantee of search improvement.

The barman result has the opposite character.  Its unary potential is weak
relative to the remaining cost, and pairwise features capture the interaction
between the shaker, shot, and ingredient variables.  The MILP finds a much
larger initial value and a much smaller search tree.  This is precisely the
regime in which the added expressiveness pays for itself.

Two modeling choices deserve scrutiny.  First, we compared the continuous LP
relaxation with the integer MILP on barman.  The relaxation has 449 nonzero
features and reaches $h(s_0)=73$ in 13.4 s; the integer solution has 466
features and the same initial value in 11.1 s, but generates 1{,}035{,}201
states rather than 1{,}050{,}771 (a further 1.5\% reduction).  The weights
differ despite the equal initial value, which confirms that $h(s_0)$ alone is
not a sufficient predictor of search quality.  Integer weights also keep the
sidecar evaluation exact and avoid numerical drift on the GPU.

Second, we compared causal pairs with all variable pairs.  The all-pairs model
has 8{,}819 features, 314{,}004 auxiliary variables, and 24.5 s of MILP time,
against 3{,}609 features, 128{,}420 auxiliaries, and \added{11.1 s} for causal
pairs.
It again reaches $h(s_0)=73$, with no larger initial value and a 2.4$\times$
larger offline model.  We therefore use causal and goal pairs throughout.  This
is an empirical choice: a domain with strong non-causal interactions could
favor a richer pair set, but the barman ablation shows that adding all pairs is
not automatically beneficial.

To isolate hardware acceleration from heuristic quality, we ran the same
version-2 sidecars through the CPU emulation build, which executes the kernel
logic serially.  This comparison shares the parsing, the CPU duplicate
detection, and the search loop; only guard/effect evaluation and heuristic
evaluation move between GPU and CPU.  Table~\ref{tab:emu} reports median
wall-clock times of three runs, including start-up, search, and plan output.
For very small state spaces the fixed GPU cost dominates and emulation is about
10\% faster.  The GPU pulls ahead at 418{,}200 states (1.5$\times$) and on
barman (4.4$\times$).  The state reductions are therefore properties of the
heuristic, not of GPU-only semantics, while GPU acceleration pays off once the
search is large enough to amortize its start-up.

\begin{table}[!htbp]
\centering
\small
\setlength{\tabcolsep}{3.5pt}
\begin{tabular}{lrrrr}
\hline
Task & States & GPU (s) & Emu (s) & Speed-up \\
\hline
blocks (pairwise)      & 30        & 2.35 & 2.05  & 0.9$\times$ \\
logistics (pairwise)   & 1{,}701  & 2.38 & 2.07  & 0.9$\times$ \\
elevators (blind)      & 418{,}200 & 3.03 & 4.45  & 1.5$\times$ \\
elevators (pairwise)   & 8{,}560   & 2.51 & 2.14  & 0.9$\times$ \\
barman (pairwise)      & 1{,}035{,}201 & 3.32 & 14.51 & 4.4$\times$ \\
\hline
\end{tabular}
\caption{GPU versus serial emulation with the same kernel logic.  Times are
medians of three runs; speed-up is GPU over emulation.  Small searches expose
fixed GPU start-up cost, while larger searches amortize it.
\added{Blocks (pairwise) generates 30 states here and 48 in
Table~\ref{tab:results}; see the remark on blocks above.}}
\label{tab:emu}
\end{table}

\begin{addedtext}
\subsection{Core Results}
We first reran the domains of Table~\ref{tab:results}, problems 1--3 of each
(45 tasks), with blind, unary and pairwise CuDPlan and with Fast Downward's
blind search and LM-cut \citep{helmert2009landmarks}, 300 s per run.  The
barman numbers of Table~\ref{tab:results} are reproduced exactly: 6{,}158{,}330,
5{,}645{,}887 and 1{,}035{,}201 generated states, with $h(s_0)=36$ and $73$.
All optimal planners agree on every cost, and every plan validates.  On the 40
tasks that all three CuDPlan configurations solve, blind search generates
106{,}583{,}521 states, unary 22{,}423{,}974 and pairwise 3{,}975{,}887:
pairwise is 82.3\% below unary and 96.3\% below blind.  The pairwise MILP times
out on tidybot (as above) and on freecell problems 2 and 3.

On this machine the barman pairwise search takes 1.23 s (median of three runs),
against 51.4 s for Fast Downward with LM-cut on the same task.  On this machine CuDPlan generates a median of 1.75 million states per second on
the solved tasks taking at least one second, with a maximum of 2.7 million.

\citet{fritzsche2026gpu} evaluate a GPU-accelerated $h^2$ heuristic inside Fast
Downward on 1{,}827 tasks from 65 domains and publish per-task results.  Their
design is complementary to ours: the expensive heuristic runs on the GPU while
search stays on the CPU, whereas CuDPlan moves successor generation to the GPU
and pairs it with cheap potentials.  We ran CuDPlan with blind search and with
unary potentials, and Fast Downward with LM-cut, on the same suite with 300 s
per run.  The closed list has $2^{27}$ entries, which keeps CuDPlan's memory in
the range of the 8 GB given to their planners; Fast Downward also had 8 GB.  To
keep the run within two 12-hour sessions, our runner skips the remaining
problems of a domain for a planner after two consecutive failures; skipped tasks
count as unsolved, so our coverage figures are lower bounds.  The comparison
times for their planners come from their published data
\citep{fritzsche2026gpu}.

Table~\ref{tab:coverage} gives the totals.  With unary potentials CuDPlan
solves 668 tasks in five minutes, against 564 for the best GPU $h^2$
configuration in five minutes and 642 in thirty.  LM-cut solves more: 820 in
our run, 886 in theirs at five minutes (our lower figure reflects the skip rule)
and 968 at thirty minutes.  Every CuDPlan plan validates except on three
data-network tasks that VAL cannot parse; there, the costs agree with LM-cut.

\begin{table}[!htbp]
\centering
\small
\setlength{\tabcolsep}{4pt}
\begin{tabular}{lrr}
\hline
Planner & 300 s & 1800 s \\
\hline
CuDPlan, blind (ours) & 588 & --- \\
CuDPlan, unary potentials (ours) & 668 & --- \\
GPU $h^2$, batched \citep{fritzsche2026gpu} & 564 & 642 \\
LM-cut (our run) & 820 & --- \\
LM-cut \citep{fritzsche2026gpu} & 886 & 968 \\
Portfolio: CuDPlan 5 min, then LM-cut & --- & 1{,}003 \\
\hline
\end{tabular}
\caption{Tasks solved out of 1{,}827 (65 domains).  Our runs are lower bounds
(a planner skips the rest of a domain after two consecutive failures).  The
portfolio row combines our CuDPlan times with the published LM-cut times of
\citet{fritzsche2026gpu}.}
\label{tab:coverage}
\end{table}

Per domain, the picture is mixed.  CuDPlan is strongest where states are cheap
to expand and the reachable space is large but regular: on tetris it solves all
17 tasks, against 3 for our LM-cut run and 6 for theirs at thirty minutes; on
tidybot-opt11 it solves 14, where the GPU $h^2$ planner solves none; on snake 10
(LM-cut: 1 in our run, 7 in theirs at thirty minutes); and on
pipesworld-notankage 23 (LM-cut: 14 and 18).  The GPU $h^2$ planner is clearly
better on domains where $h^2$ prunes strongly, for example transport-opt08
(11 against 4), mystery (11 against 3), mprime (19 against 10) and
data-network (11 against 3).  Figure~\ref{fig:socs} compares runtimes per task
with the GPU $h^2$ planner and calibrates the two machines with LM-cut, the one
planner run on both; Figure~\ref{fig:cross} compares CuDPlan with LM-cut on our
machine.

\begin{figure*}[!htbp]
\centering
\includegraphics[width=0.85\textwidth]{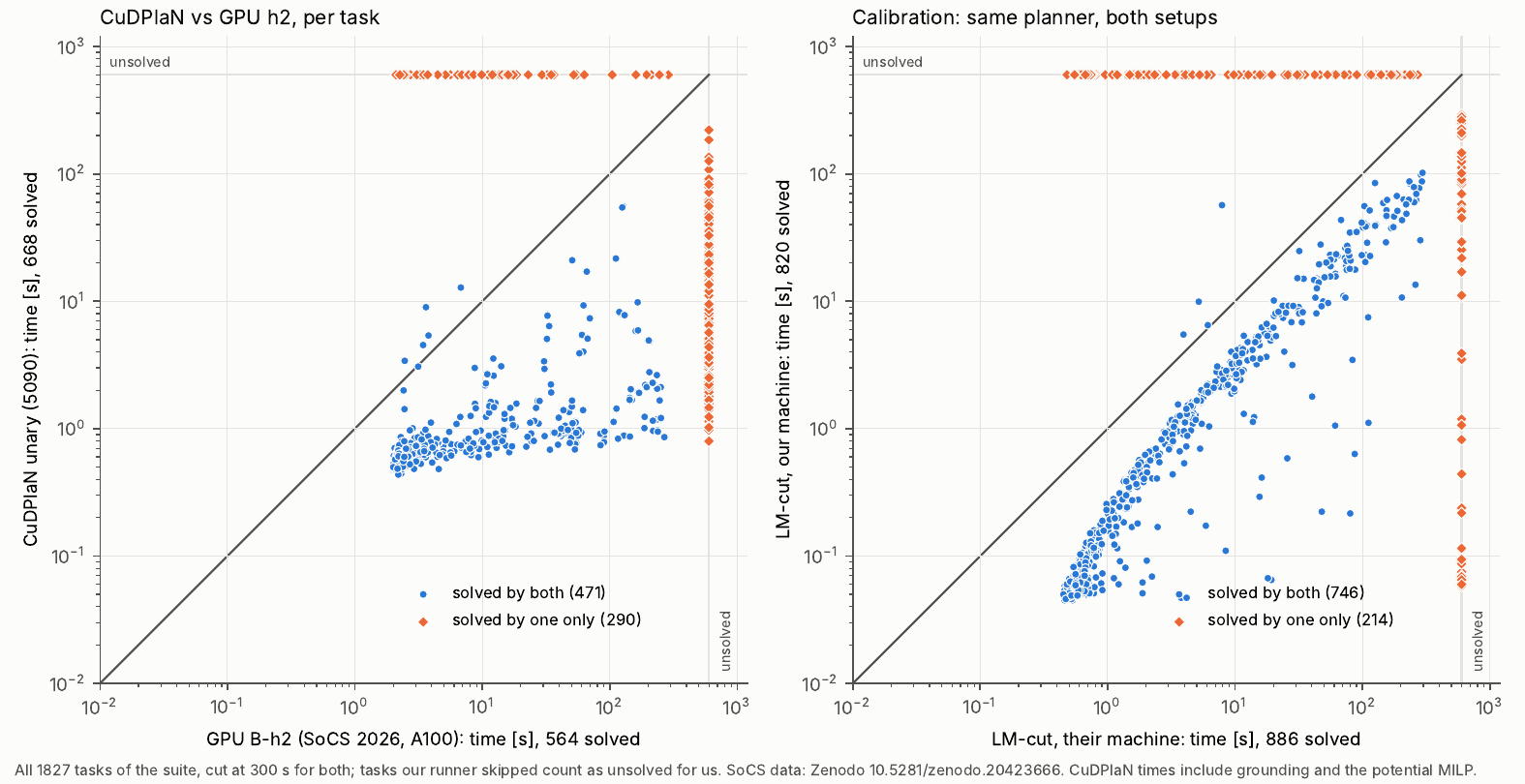}
\caption{Left: per-task runtime of CuDPlan with unary potentials against the
batched GPU $h^2$ planner of \citet{fritzsche2026gpu}, from their published
data.  Right: LM-cut on our machine against LM-cut on theirs; our machine is
faster per task, which should be kept in mind when reading the left plot.}
\label{fig:socs}
\end{figure*}

\begin{figure*}[!htbp]
\centering
\includegraphics[width=\textwidth]{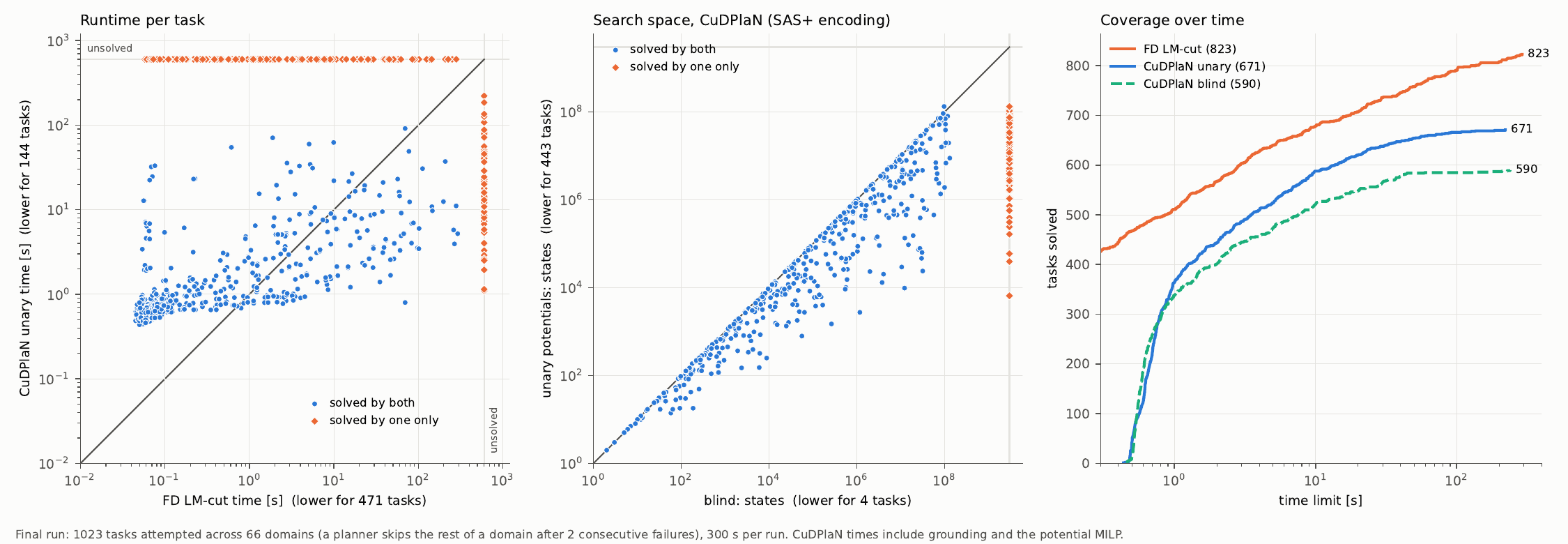}
\caption{CuDPlan against Fast Downward on our machine, 300 s per run.  Left:
runtime per task, CuDPlan with unary potentials against LM-cut.  Middle:
generated states, unary potentials against blind search.  Right: coverage as a
function of the time limit.  This figure covers all 1{,}023 tasks we attempted
in 66 domains (the suite plus quantum-layout), hence 671 and 823 rather than
668 and 820.}
\label{fig:cross}
\end{figure*}

Most of CuDPlan's failures in the coverage run are memory-outs rather than
timeouts (124 against 9 for unary potentials): the closed list fills before
the time limit.  We therefore reran the 125 tasks on which unary CuDPlan had run
out of memory, sizing the closed list per task up to a 32 GiB budget, again
with 300 s.  Of these, 52 are now solved, 46 time out, and 27 still exhaust
memory.  Coverage over all 1{,}023 attempted tasks rises from 671 to 723
(Figure~\ref{fig:memory}); on the 1{,}827-task suite, which does not include
quantum-layout, from 668 to 719.  The largest search, logistics00
probLOGISTICS-10-1, generates 1.04 billion states in 295 s.  The default
closed list holds $2^{27}\approx 134$ million states; many of the newly solved
tasks need several hundred million (Figure~\ref{fig:states}).  We report this
run separately from Table~\ref{tab:coverage} because it no longer matches the
memory of the planners we compare with.

\begin{figure}[!htbp]
\centering
\includegraphics[width=\columnwidth]{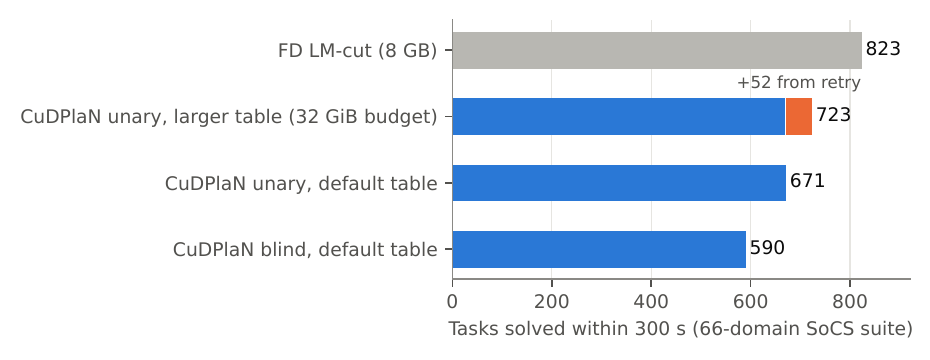}
\caption{Coverage within 300 s on the 66 domains we ran (the SoCS suite plus
quantum-layout).  The orange segment marks the 52 tasks solved after enlarging
the closed list.}
\label{fig:memory}
\end{figure}

\begin{figure}[!htbp]
\centering
\includegraphics[width=\columnwidth]{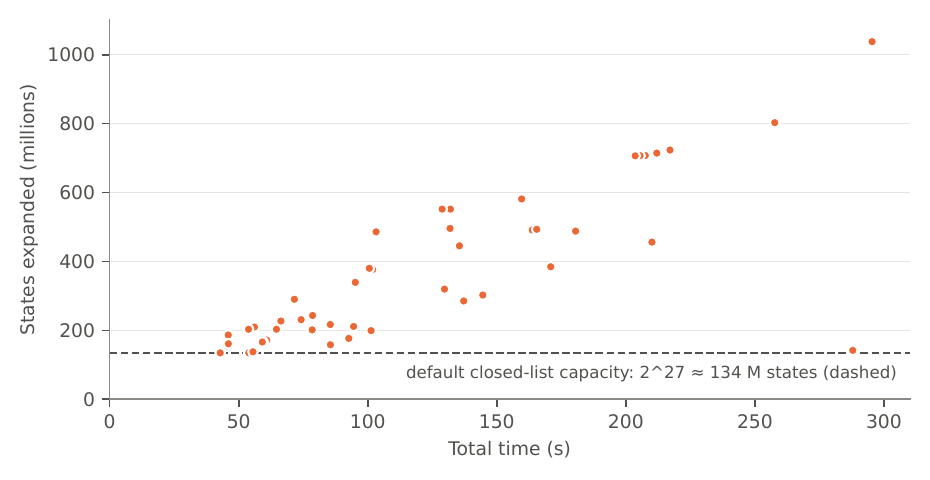}
\caption{Tasks solved only with the enlarged closed list: generated states
against runtime.  The dashed line is the default capacity of $2^{27}$ states.}
\label{fig:states}
\end{figure}

CuDPlan and LM-cut fail on different tasks: CuDPlan solves 44 tasks in five
minutes that LM-cut does not solve in thirty.  A sequential portfolio exploits
this.  It runs CuDPlan with unary potentials for up to five minutes and hands
over to Fast Downward with LM-cut for the remaining twenty-five; the handover
happens earlier when CuDPlan fills its closed list.  Combining our CuDPlan
times with the published LM-cut times \citep{fritzsche2026gpu}, the portfolio
solves 1{,}003 of the 1{,}827 tasks within thirty minutes, against 968 for
LM-cut alone and 642 for the GPU $h^2$ planner (Figure~\ref{fig:portfolio}).
It loses nine tasks that LM-cut solves only after more than 25 minutes.  The
two halves were timed on different machines; a same-machine version needs only
52 additional LM-cut runs of 25 minutes.

\begin{figure}[!htbp]
\centering
\includegraphics[width=0.8\columnwidth]{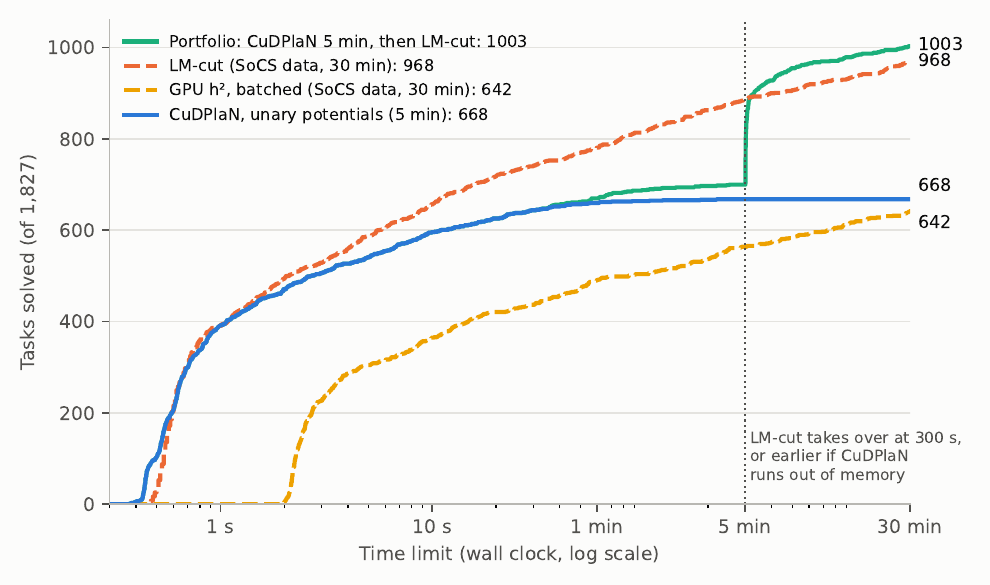}
\caption{Coverage over time on the 1{,}827 tasks.  The portfolio runs CuDPlan
with unary potentials for up to 5 minutes, then LM-cut; LM-cut and GPU $h^2$
times are from the published data of \citet{fritzsche2026gpu}.}
\label{fig:portfolio}
\end{figure}

With an enlarged closed list, CuDPlan with unary potentials solves barman
pfile02-005, generating 166.3 million states in 59.5 s with cost 121, a task
Fast Downward with LM-cut does not solve within 300 s in our coverage run.
With unary potentials and a closed list of $2^{28}$ entries, CuDPlan solves
tetris p01-10 (121.8 million states, 44 s, median of three runs), where Fast
Downward with LM-cut times out at 600 s and blind Fast Downward runs out of
memory.  The opposite also occurs: on
parking pfile03-011 CuDPlan needs 227 million states and 66 s, while LM-cut
expands 643 states in 1.2 s, and on floortile LM-cut is eight to nine times faster than unary CuDPlan.  Brute-force throughput helps most when the heuristic of the
competitor is weak relative to the size of the space.

On the solved tasks of this set taking at least one second, the GPU guard
check and effect application take a median of 5.4\% and 2.1\% of wall-clock
time, while CPU
duplicate detection takes 46.1\% (51.0\% on the larger tasks).  The bottleneck of
the revived planner is therefore the CPU hash table, not the GPU.  Moving
duplicate detection to the GPU, which the original code supports as an
experimental option, is the most promising route to further speed.

\subsection{Transfer Results}
The Metal path of the portable kernels transfers the planner to Apple's
unified-memory GPUs.  On the barman task it solves in 9.5 s against 45.0 s
for the CPU emulation (a 4.8$\times$ speedup); on sub-second tasks the two
stay within 0.1 s of each other, where the fixed launch overheads dominate.
The split design---batched expansion on the GPU, duplicate detection in a
CPU hash table---suits unified-memory architectures particularly well: the
CPU and the GPU share one address space, so the layer round-trips need no
copies between RAM and VRAM.  The property is not Apple-specific: NVIDIA's
DGX and RTX Spark systems also provide unified memory, which makes the
planner available across consumer and data-center hardware.  The same
portability makes CuDPlan a natural component for the unified planning
framework of AIPlan4EU (\url{https://www.aiplan4eu-project.eu/}), whose goal
is to bring planning tools to a broad user base.
\end{addedtext}

\section{Discussion and Demarcation}
The evaluation uses one instance per domain, not the full IPC suite, so the
per-domain numbers should be read as case studies rather than benchmark
averages.  The barman task dominates the aggregate, and the thirteen-instance
total is dominated by elevators; we therefore report both per-instance and
aggregate figures.  The binary encoding is shared by all three configurations,
which makes the unary-versus-pairwise comparison controlled, but it is not the
one-hot encoding used by some earlier atomic experiments.  The pairwise feature
set is one reasonable choice (causal and goal pairs, weight bound 1000), not an
exhaustive search over feature families.  Finally, the MILP is solved offline;
a different objective or a time-limited solve could select a different
potential with a different search profile.  The plan-validity and equal-cost
checks guard against implementation errors, not against these modeling choices.
\added{For the coverage comparison, the planners ran on different hardware: the
LM-cut calibration in Figure~\ref{fig:socs} shows our machine to be faster per
task, so part of CuDPlan's advantage over the published GPU $h^2$ times is due
to the CPU.  Our own coverage figures are lower bounds because of the skip rule,
and the portfolio combines times from two machines.}

\section{Conclusion}
We have revived CuDPlan and extended it, without breaking its atomic
heuristic path, with GPU-evaluated conjunctive potentials.  The new sidecar is
small, versioned, and independent of the offline solver.  Pairwise features
produce an 81.7\% state reduction on barman and improve ten of thirteen smaller
instances, but they are not a universal replacement for unary features: three
instances regress, and the aggregate can hide that.  The practical lesson is to
treat the feature dimension as a tuning knob selected per domain, not as a
monotone quality scale.
\added{On a 1{,}827-task suite, the revived planner with cheap unary potentials
solves more tasks in five minutes than the published results of a
GPU-accelerated $h^2$ planner in thirty (on different hardware), and as the
first stage of a portfolio it lifts LM-cut from 968 to 1{,}003 solved
tasks.}
Future work includes cheaper feature selection,
MILP decomposition for large tasks such as tidybot, and completing the Slang
port of the version-2 kernel\added{, duplicate detection on the GPU, and
external-memory search to go beyond the closed list held in RAM}.

\begin{addedtext}
\section{Acknowledgments}
Thanks to Damian Sulewski for providing sources of CuDPlan.
Most of the porting, the experiment tooling and the drafting were done with
an LLM coding assistant (Claude, Anthropic); all results were checked by the
authors.
\end{addedtext}

\clearpage
\bibliography{references}

\end{document}